%% file: kddSRR.tex
\documentclass{article}

\usepackage{amsfonts}
\usepackage{bm}
\usepackage{times}

\usepackage{epsfig}
\usepackage{graphicx}

\usepackage{amsmath}
\usepackage{amssymb}
\usepackage{multirow}
\usepackage{algorithm}
\usepackage{algorithmic}
\usepackage{url}

\usepackage{moresize}

\newcounter{counter_theorem}
\newcounter{counter_lemma}

\newtheorem{theorem}[counter_theorem]{Theorem}
\newtheorem{lemma}[counter_lemma]{Lemma}

\begin{document}

\title{Successive Ray Refinement and Its Application to Coordinate Descent for LASSO}

\author{
Jun Liu\thanks{Corresponding author. E-mail address: junliu.nt@gmail.com.}\\
SAS Institute Inc.
\and
Zheng Zhao\thanks{This work was done when Zheng Zhao was with SAS.}\\
Google Inc.\\
\and
Ruiwen Zhang \\
SAS Institute Inc.
}

\maketitle

\begin{abstract}
Coordinate descent is one of the most popular approaches for solving Lasso and its extensions due to its simplicity and efficiency.
When applying coordinate descent to solving Lasso, we update one coordinate at a time while fixing the remaining coordinates.
Such an update, which is usually easy to compute, greedily decreases the objective function value.
In this paper, we aim to improve its computational efficiency by reducing the number of coordinate descent iterations.
To this end, we propose a novel technique called Successive Ray Refinement (SRR). 
SRR makes use of the following ray continuation property on the successive iterations:
for a particular coordinate, the value obtained in the next iteration almost always
lies on a ray that starts at its previous iteration and passes through the current iteration. 
Motivated by this ray-continuation property, we propose that coordinate descent be performed not directly on the previous iteration but on
a refined search point that has the following properties: on one hand, it lies on a ray that
starts at a history solution and passes through the previous iteration, and
on the other hand, it achieves the minimum objective function value among all the points on the ray. 
We propose two schemes for defining the search point and
show that the refined search point can be efficiently obtained. 
Empirical results for real and synthetic data sets show that the proposed SRR can significantly reduce the number of coordinate descent iterations,
especially for small Lasso regularization parameters.
\end{abstract}


\section{Introduction}

\input{introduction}

\section{Coordinate Descent For Lasso}\label{s:traditional:lasso}

\input{coordinateDescent}

\section{Successive Ray Refinement}\label{s:sfg}

\input{srr}

\section{Efficient Refinement Factor\\ Computation}\label{s:one:d:search}

\input{alphaCompute}

\section{An Eigenvalue Analysis on \\the Proposed SRR}\label{s:sfg:eigen:analysis}

\input{justification}

\section{Related Work}\label{s:related}

\input{relatedWork}

\section{Experiments}\label{s:experiment}

\input{experiment}

\section{Conclusion}\label{s:conclusion}

\input{conclusion}


\end{document}

%% file: introduction.tex
Lasso~\cite{Tibshirani:Lasso:1996}  is an effective technique for analyzing high-dimensional data. 
It has been applied successfully in various areas, such as machine learning, signal processing, image processing,
medical imaging, and so on. 
Let $X =[\mathbf x_1, \mathbf x_2, \ldots, \mathbf x_p] \in \mathcal{R}^{n \times p}$ denote the data matrix composed of $n$ samples with $p$ variables, and let
$\mathbf y \in \mathcal{R}^{n \times 1}$ be the response vector.
In Lasso, we compute the $\bm \beta$ that optimizes
\begin{equation}\label{eq:lasso:problem}
  \min_{\bm \beta} f(\bm \beta) = \frac{1}{2} \| X \bm \beta - \mathbf y\|_2^2 + \lambda \|\bm \beta\|_1,
\end{equation}
where the first term measures the discrepancy between the prediction and the response and 
the second term controls the sparsity of $\bm \beta$ with $\ell_1$ regularization.
The regularization parameter $\lambda$ is nonnegative, and a larger $\lambda$ usually leads to 
a sparser solution.

Researchers have developed many approaches for solving Lasso in Equation~\eqref{eq:lasso:problem}.
Least Angle Regression (LARS)~\cite{Efron:LAR:2004} is one of the most well-known homotopy approaches for Lasso.
LARS adds or drops one variable at a time, generating a piecewise linear solution path for Lasso.
Unlike LARS, other approaches usually solve Equation~\eqref{eq:lasso:problem} according to some
prespecified regularization parameters. These methods include the coordinate descent method~\cite{Friedman:coordinate:2010,Yuan:2012:l1:logistic:jmlr},
the gradient descent method~\cite{beck:2009:fast,wright:2009:sparse}, the interior-point method~\cite{Koh:sparse:logistic:2007}, 
the stochastic method~\cite{Shai:2009:icml:stochastic:method}, and so on.
Among these approaches, coordinate descent is one of the most popular approaches due to its simplicity and efficiency.
When applying coordinate descent to Lasso, we update one coordinate at a time while fixing the remaining coordinates.
This type of update, which is easy to compute, can effectively decrease the objective function value in a greedy way.

To improve the efficiency of optimizing the Lasso problem in Equation~\eqref{eq:lasso:problem},
the screening technique has been extensively studied in~\cite{Ghaoui:2012,liu_j:14,Ogawa2013,Tibshirani:2012,Wang:2012:report,Xiang:2011}.
Screening 1) identifies and removes the variables that have zero entries
in the solution $\bm \beta$ and 2) solves Equation~\eqref{eq:lasso:problem} by using only the kept variables.
When one is able to discard the variables that have zero entries in the final solution $\bm \beta$ and
identify the signs of the nonzero entries, the Lasso problem in Equation~\eqref{eq:lasso:problem} becomes
a standard quadratic programming problem. However, it is usually very hard to identify all the zero entries,
especially when the regularization parameter is small. 
In addition, the computational cost of Lasso usually increases as the the regularization parameter decreases.
The computational cost increase motivates us to come up with an approach that can accelerate the computation of Lasso for small regularization parameters.

\begin{figure}
  \centering
	\includegraphics[width=0.7\columnwidth]{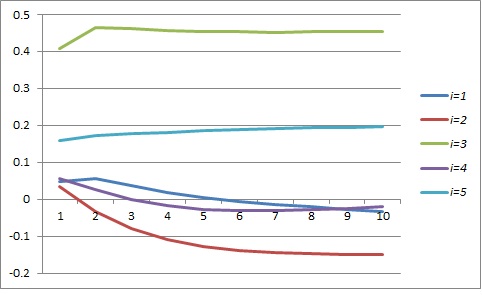}\\
	(a)\\
	\includegraphics[width=0.7\columnwidth]{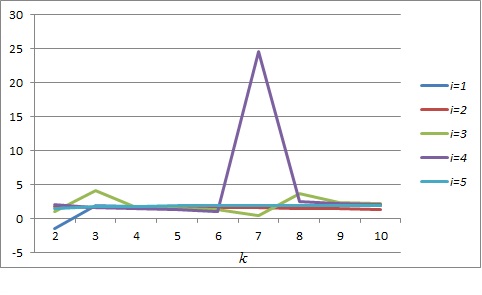}\\
	(b)
	\caption{Illustration of the iterations of coordinate descent. For both plots, the x-axis corresponds to the iteration number $k$.
	The y-axis of plot (a) denotes $\beta_i^k$, the value of the $i$th coordinate in the $k$th iteration.
	The y-axis of plot (b) denotes $\alpha_i^k$, which is computed using the equation $\beta_i^{k+1} =\alpha^k_i \beta_i^{k-1} + (1-\alpha_i^k) \beta_i^k$.
	Ray-continuation property: for a given coordinate $i$, the value obtained in the next iteration denoted by $\beta_i^{k+1}$ almost always lies on a ray that starts at its previous iteration, $ \beta_i^{k-1}$, and passes through the current iteration, $\alpha_i^k$.
  For numerical details of plot (a) and plot (b), see Table~\ref{table:iteration:traditiona:cd} and Table~\ref{table:iteration:traditiona:alpha}.
	}	\label{fig:cd:solution:alpha}
\end{figure}

In this paper, we aim to improve the computational efficiency of coordinate descent by reducing its iterations.
To this end, we propose a novel technique called Successive Ray Refinement (SRR). 
Our proposed SRR is motivated by an interesting ray-continuation property on the coordinate descent iterations:
for a given coordinate, the value obtained in the next iteration almost always
lies on a ray that starts at its previous iteration and passes through the current iteration. 
Figure~\ref{fig:cd:solution:alpha} illustrates the ray-continuation property by using the data specified in Section~\ref{s:traditional:lasso}.
Motivated by this ray-continuation property, we propose that coordinate descent be performed not directly on the previous iteration but on
a refined search point that has the following properties: on one hand, the search point lies on a ray
that starts at a history solution and passes through the previous iteration, and
on the other hand, the search point achieves the minimum objective function value among all the points on the ray.
We propose two schemes for defining the search point, and
we show that the refined search point can be efficiently computed. 
Experimental results on both synthetic and real data sets demonstrate that the proposed SRR can greatly accelerate the convergence of coordinate descent 
for Lasso, especially when the regularization parameter is small.

\vspace{0.1in}

\noindent \textbf{Organization } The rest of this paper is organized as follows. 
We introduce the traditional coordinate descent for Lasso and present the ray-continuation property that motivates this paper in Section~\ref{s:traditional:lasso},
propose the SRR technique in Section~\ref{s:sfg}, discuss the efficient computation of the refinement factor that is used in SRR in Section~\ref{s:one:d:search},
conduct an eigenvalue analysis on the proposed SRR in Section~\ref{s:sfg:eigen:analysis}, and compare SRR with related work in Section~\ref{s:related}.
We report experimental results on both synthetic and real data sets in Section~\ref{s:experiment}, and we conclude this paper in Section~\ref{s:conclusion}.

\vspace{0.1in}

\noindent \textbf{Notations } Throughout this paper, scalars are denoted by italic letters and vectors by bold face letters. 
Let $\|\cdot\|_1$ denote the $\ell_1$ norm, let $\|\cdot\|_2$ denote the Euclidean norm, and let $\|\cdot\|_{\infty}$ denote the infinity norm.
Let $\langle \mathbf x, \mathbf y \rangle$ denote the inner product between $\mathbf x$ and $\mathbf y$. 
Let a superscript denote the iteration number, and let a subscript denote
the index of the variable or coordinate. We assume that $X$ does not contain a zero column; that is, $\|\mathbf x_i\|_2 \neq 0, \forall i$.

%% file: coordinateDescent.tex
In this section, we first review the coordinate descent method for solving Lasso, and then analyze the adjacent iterations to motivate the proposed SRR technique.

Let $\beta_i^k$ denote the $i$th element of $\bm \beta$, which is obtained at the $k$th iteration of coordinate descent.
In coordinate descent, we compute
$\beta_i^k$ while fixing $\beta_j= \beta_j^k, 1 \le j < i $, and $\beta_j= \beta_j^{k-1}, i <  j \le p$. Specifically, $\beta_i^k$ 
is computed as the minimizer to the following univariate optimization problem:
\begin{equation*}
   \beta_i^k= \arg \min_{\beta} f([ \beta_1^k,\ldots,\beta_{i-1}^k,\beta,\beta_{i+1}^{k-1},\ldots,\beta_{p}^{k-1}   ]^T).
\end{equation*}
It can be computed in a closed form as:
\begin{equation}\label{eq:beta:i:kplus1}
\beta_i^k = \frac{ S(\mathbf x_i^T \mathbf y - \sum_{j<i} \mathbf x_i^T \mathbf x_j \beta_j^k - \sum_{j>i} \mathbf x_i^T \mathbf x_j \beta_j^{k-1} , \lambda) }{\|\mathbf x_i\|_2^2},
\end{equation}
where $S(\cdot, \cdot)$ is the shrinkage function
\begin{equation}\label{eq:shrinkage:operator}
  S(x, \lambda) =
	\left\{
	\begin{array}{cc} 
x -\lambda & x > \lambda \\
x +\lambda & x < -\lambda \\
0    & |x | \le\lambda. 
\end{array}		\right.
\end{equation}
Let 
\begin{equation}\label{eq:residual}
   \mathbf r_i^k = \mathbf y - X [ \beta_1^k,\ldots,\beta_{i-1}^k,\beta_i^k,\beta_{i+1}^{k-1},\ldots,\beta_{p}^{k-1}  ]^T
\end{equation}
denote the residual obtained after updating $\beta_i^{k-1}$ to $\beta_i^k$. 
With Equation~\eqref{eq:residual}, we can rewrite Equation~\eqref{eq:beta:i:kplus1} as
\begin{equation}\label{eq:beta:i:kplus1:residual}
\beta_i^k =   S( \beta_i^{k-1} + \frac{\mathbf x_i^T \mathbf r_{i-1}^k}{\|\mathbf x_i\|_2^2} , \frac{\lambda} {\|\mathbf x_i\|_2^2}).
\end{equation}
In addition, with the updated $\beta_i^k$, we can update the residual from $\mathbf r_{i-1}^k$ to $\mathbf r_i^k$ as
\begin{equation}\label{eq:residual:udpated}
   \mathbf r_i^k = \mathbf r_{i-1}^k + \mathbf x_i (\beta_i^{k-1} - \beta_i^k).
\end{equation}

Algorithm~\ref{algorithm:traditional} illustrates solving Lasso via coordinate descent. 
Since the non-smooth $\ell_1$ penalty in Equation~\eqref{eq:lasso:problem} is separable,
the algorithm is guaranteed to converge~\cite{Tseng01convergenceof}.

\begin{algorithm}   
\caption{Coordinate Descent for Lasso}        
\label{algorithm:traditional}
\begin{algorithmic}[1] 
    \REQUIRE $X$, $\mathbf y$, $\lambda$
    \ENSURE $\bm \beta^k$
    \STATE $k=0$, $\bm \beta^0 = \mathbf 0$, $\mathbf r^0 = \mathbf y$
    \REPEAT
				\STATE Set $k=k+1$, $\mathbf r^k = \mathbf r^{k-1}$
        \FOR{$i=1$  to $p$}
				   \STATE Compute $\beta_i^k =   S( \beta_i^{k-1} + \frac{\mathbf x_i^T \mathbf r^k}{\|\mathbf x_i\|_2^2} , \frac{\lambda} {\|\mathbf x_i\|_2^2})$
				   \STATE Update residual $\mathbf r^k = \mathbf r^k + \mathbf x_i (\beta_i^{k-1} - \beta_i^k)$
				\ENDFOR
    \UNTIL{convergence criterion satisfied}
\end{algorithmic}
\end{algorithm}

\begin{table}
\caption{Applying coordinate descent in Algorithm~\ref{algorithm:traditional} to solving Lasso in Equation~\eqref{eq:lasso:problem} with $\lambda=0$.
Since $\lambda=0$ and $X$ is invertible, the optimal function value is 0.}
\begin{center}
\begin{tabular}{ccccccc}
  \hline
  $k$ & $\beta_1^k$ & $\beta_2^k$ & $\beta_3^k$ & $\beta_4^k$ & $\beta_5^k$ & $f(\bm \beta^k)$\\
  \hline\hline
1
 & 
0.048912 & 0.034041 & 0.407960 & 0.055687 & 0.160413
 & 
0.052449
\\ \hline
2
 & 
0.057182 & -0.033692 & 0.465254 & 0.027810 & 0.171740
 & 
0.017591
\\ \hline
3
 & 
0.036909 & -0.079955 & 0.463604 & -0.000612 & 0.177708
 & 
0.008085
\\ \hline
4
 & 
0.019050 & -0.108954 & 0.458440 & -0.017618 & 0.182115
 & 
0.003933
\\ \hline
5
 & 
0.005418 & -0.126712 & 0.455218 & -0.026135 & 0.185698
 & 
0.002304
\\ \hline
6
 & 
-0.005122 & -0.137403 & 0.453694 & -0.029295 & 0.188740
 & 
0.001653
\\ \hline
7
 & 
-0.013567 & -0.143688 & 0.453262 & -0.029210 & 0.191398
 & 
0.001358
\\ \hline
8
 & 
-0.020585 & -0.147239 & 0.453491 & -0.027216 & 0.193771
 & 
0.001187
\\ \hline
9
 & 
-0.026604 & -0.149101 & 0.454104 & -0.024144 & 0.195923
 & 
0.001060
\\ \hline
10
 & 
-0.031899 & -0.149927 & 0.454929 & -0.020507 & 0.197895
 & 
0.000950
\\ \hline
...
\\ \hline
28
 & 
-0.081090 & -0.142355 & 0.467964 & 0.030835 & 0.217715
 & 
0.000106
\\ \hline
29
 & 
-0.082490 & -0.142044 & 0.468368 & 0.032406 & 0.218288
 & 
0.000093
\\ \hline
30
 & 
-0.083806 & -0.141752 & 0.468749 & 0.033883 & 0.218827
 & 
0.000082
\\ \hline
...
\\ \hline
100
 & 
-0.103999 & -0.137267 & 0.474584 & 0.056543 & 0.227099
 & 
1.3349e-08
\\ \hline
101
 & 
-0.104015 & -0.137264 & 0.474589 & 0.056561 & 0.227105
 & 
1.1785e-08
\\ \hline
102
 & 
-0.104030 & -0.137261 & 0.474593 & 0.056577 & 0.227111
 & 
1.0403e-08
\\ \hline
103
 & 
-0.104044 & -0.137258 & 0.474597 & 0.056593 & 0.227117
 & 
9.1839e-09
\\ \hline
104
 & 
-0.104057 & -0.137255 & 0.474601 & 0.056608 & 0.227122
 & 
8.1074e-09
\\ \hline
105
 & 
-0.104069 & -0.137252 & 0.474604 & 0.056621 & 0.227127
 & 
7.1571e-09
\\ \hline
\end{tabular}
\end{center}
\vspace{-0.1in}
\label{table:iteration:traditiona:cd}
\end{table}

We demonstrate Algorithm~\ref{algorithm:traditional} 
using the following randomly generated $X$ and $\mathbf y$:

\begin{equation}\label{eq:X}
X = \left[
\begin{array}{ccccc}
-0.204708 & 0.478943 & -0.519439 & -0.555730 & 1.965781
\\
1.393406 & 0.092908 & 0.281746 & 0.769023 & 1.246435
\\
1.007189 & -1.296221 & 0.274992 & 0.228913 & 1.352917
\\
0.886429 & -2.001637 & -0.371843 & 1.669025 & -0.438570
\\
-0.539741 & 0.476985 & 3.248944 & -1.021228 & -0.577087
\end{array}
		\right],
\end{equation}
\begin{equation}\label{eq:y}
  \mathbf y = [0.124121,  0.302614,  0.523772,  0.000940,  1.343810]^T.
\end{equation}

We show the iterations of coordinate descent for Lasso with $\lambda=0$ in
Table~\ref{table:iteration:traditiona:cd} and 
Figure~\ref{fig:cd:solution:alpha} (a).
We set $\lambda=0$ to facilitate the eigenvalue analysis
in Section~\ref{s:sfg:eigen:analysis}.
Note that the results reported here also generalize to Lasso,
because if we know the sign of the optimal solution $\bm \beta^*$, the 
nonzero entries of $\bm \beta^*$ can be solved by the following equivalent convex smooth problem:
\begin{equation}\label{eq:lasso:problem:lambda:0}
  \min_{\bm \beta} \frac{1}{2} \|  \sum_{i: s_i \neq 0} \mathbf x_i \beta_i- \mathbf y\|_2^2 + \lambda \sum_{i: s_i \neq 0} \beta_i s_i,
\end{equation}
where
$s_i =0$ if $\beta^*_i =0$, $s_i =1$ if $\beta^*_i >0$, and $s_i =-1$ if $\beta^*_i <0$.

It can be observed from the results in Table~\ref{table:iteration:traditiona:cd} and Figure~\ref{fig:cd:solution:alpha} (a)
that we can obtain an approximate solution with a small objective function 
value within a few iterations. However, achieving a solution with high precision takes quite a few iterations for this example. More interestingly, 
for a particular coordinate, the value obtained in the next iteration almost always
lies on a ray that starts at its previous iteration and passes through the current iteration. 
To show this, we compute $\alpha^k_i$ that satisfies the following equation:
\begin{equation}
  \beta_i^{k+1} =\alpha^k_i \beta_i^{k-1} + (1-\alpha_i^k) \beta_i^k.
\end{equation}
Table~\ref{table:iteration:traditiona:alpha} and Figure~\ref{fig:cd:solution:alpha} (b) show the values of $\alpha_i^k$ for different iterations. 
It can be observed that the values of $\alpha_i^k$ are almost always positive except $\alpha_1^2$ for this example. 
In addition, most of the values of $\alpha_i^k$ are larger than 1.
We tried quite a few synthetic data and observed a similar phenomenon.

\begin{table}
\caption{Illustration of the ray-continuation property
$\beta_i^{k+1} =\alpha^k_i \beta_i^{k-1} + (1-\alpha_i^k) \beta_i^k$ based on
the results obtained in Table~\ref{table:iteration:traditiona:cd}. All the $\alpha_i^k$ are positive except $\alpha_1^2$.}
\begin{center}
\begin{tabular}{cccccc}
  \hline
  $k$ & $\alpha_1^k$ & $\alpha_2^k$ & $\alpha_3^k$ & $\alpha_4^k$ & $\alpha_5^k$ \\
  \hline\hline
2
 & 
-1.451503 & 1.683019 & 0.971191 & 2.019562 & 1.526899
\\ \hline
3
 & 
1.880924 & 1.626844 & 4.128185 & 1.598324 & 1.738268
\\ \hline
4
 & 
1.763341 & 1.612353 & 1.624041 & 1.500861 & 1.813212
\\ \hline
5
 & 
1.773143 & 1.602012 & 1.473119 & 1.370962 & 1.849048
\\ \hline
6
 & 
1.801288 & 1.587906 & 1.283008 & 0.973180 & 1.873809
\\ \hline
7
 & 
1.830983 & 1.564952 & 0.469896 & 24.529117 & 1.892594
\\ \hline
8
 & 
1.857679 & 1.524402 & 3.679312 & 2.540794 & 1.906640
\\ \hline
9
 & 
1.879781 & 1.443912 & 2.346272 & 2.183873 & 1.916832
\\ \hline
10
 & 
1.897021 & 1.240529 & 2.128716 & 2.068764 & 1.924043
\\ \hline
...
\\ \hline
28
 & 
1.939534 & 1.939949 & 1.939646 & 1.939628 & 1.939556
\\ \hline
29
 & 
1.939545 & 1.939821 & 1.939620 & 1.939608 & 1.939560
\\ \hline
30
 & 
1.939552 & 1.939736 & 1.939602 & 1.939594 & 1.939562
\\ \hline
\end{tabular}
\end{center}
\vspace{-0.1in}
\label{table:iteration:traditiona:alpha}
\end{table}

For a particular iteration number $k$, if $\alpha_i^k=\alpha, \forall i$, we can easily
achieve $\bm \beta^{k+1} = \alpha \bm \beta^{k-1} + (1-\alpha) \bm \beta^k$ without needing to
perform any coordinate descent iteration. This motivated us 
to come up with the successive ray refinement technique 
to be discussed in the next section.

%% file: srr.tex
In the proposed SRR technique, we make use of the ray-continuation property shown in 
Figure~\ref{fig:cd:solution:alpha}, Table~\ref{table:iteration:traditiona:cd}, and Table~\ref{table:iteration:traditiona:alpha}.
Our idea is as follows: To obtain $\bm \beta^{k+1}$, we perform coordinate descent based on a refined search point 
$\mathbf s^k$ rather than on its previous solution  $\bm \beta^{k}$.
We propose setting the refined search point as:
\begin{equation}\label{eq:SGR:scheme}
   \mathbf s^k = (1-  \alpha^k)  \mathbf h^k +   \alpha^k \bm \beta^k,
\end{equation}
where $ \mathbf h^k$ is a properly chosen history solution, $\bm \beta^k$ is the current solution,
and $\alpha^k$ is an optimal refinement factor that optimizes the following univariate optimization problem:
\begin{equation}\label{eq:one:d:search}
     \alpha ^k= \arg \min_{\alpha} \{ g(\alpha) = f( (1-\alpha) \mathbf h^k + \alpha \bm \beta^k ) \}.
\end{equation}
The setting of $ \mathbf h^k$ to one of the history solutions is based on the following two considerations.
First, we aim to use the ray-continuation property to reduce the number of iterations. Second,
we need to ensure that the univariate optimization problem in Equation~\eqref{eq:one:d:search} can be efficiently computed. 
We discuss the computation of Equation~\eqref{eq:one:d:search} in Section~\ref{s:one:d:search}.

\begin{figure}[h]
  \centering
	\includegraphics[width=\columnwidth]{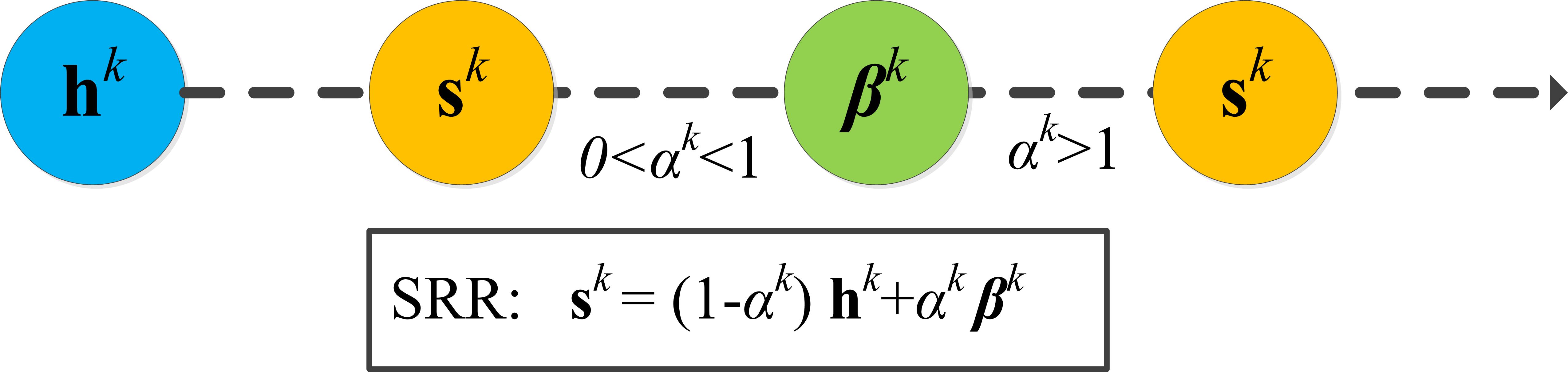}
	\caption{The proposed SRR technique. 
	The search point $\mathbf s^k$ lies on the ray that starts from a properly chosen history solution $\mathbf h^k$ and passes through
	the current solution $\bm \beta^k$, and meanwhile it achieves the minimum objective function value among all the points on the ray, optimizing Equation~\eqref{eq:one:d:search}.}
	\label{fig:sor:diagram}
\end{figure}

Figure~\ref{fig:sor:diagram} illustrates the proposed SRR technique.
When $ \alpha^k=1$, we have $\mathbf s^k=\bm \beta^k$; that is, the refined search point becomes the current solution $\bm \beta^k$.
When $ \alpha^k=0$, we have $\mathbf s^k= \mathbf h^k$; that is, the refined search point becomes the specified history solution
$\mathbf h^k$. However, our next theorem shows that $\mathbf s^k \neq \mathbf h^k$ because $\alpha^k$ is always positive. In other words,
the search point always lies on a ray that starts with the history point $\mathbf h^k$ and passes through the current solution $\bm \beta^k$.

\begin{theorem}\label{theorem:non:negative}
Assume that the history point $\mathbf h^k$ satisfies
\begin{equation}\label{eq:cd:non:increasing}
  f( \mathbf h^k ) > f(\bm \beta^k).
\end{equation}
Then, $ \alpha^k$ that minimizes Equation~\eqref{eq:one:d:search} is positive.
In addition, if $X  \mathbf h^k \neq X \bm \beta^k$, $ \alpha^k$ is unique.
\end{theorem}
\noindent \textbf{Proof }
It is easy to verify that $g(\alpha)$ is convex. Therefore, $ \alpha^k$ that minimizes Equation~\eqref{eq:one:d:search} has at least one solution.
Equation~\eqref{eq:cd:non:increasing} leads to
\begin{equation}\label{eq:g:1:0}
  g(1) < g(0).
\end{equation}
Therefore, the global refinement factor $  \alpha^k \neq 0$. 
Next, we show that $  \alpha^k$ cannot be negative. 

If $  \alpha ^k<0$, due to the convexity of $g(\alpha)$, we have
\begin{equation}
   g((1-\theta)   \alpha^k + \theta) \le (1-\theta) g(  \alpha ^k) + \theta g(1), \forall \theta \in [0,1].
\end{equation}
Setting $\theta = \frac{  \alpha^k}{  \alpha^k -1}$, we have
\begin{equation}
   g(0) \le \frac{-1}{  \alpha^k -1} g(  \alpha^k ) + \frac{  \alpha^k}{  \alpha^k -1} g(1), \forall \theta \in [0,1].
\end{equation}
Making use of Equation~\eqref{eq:g:1:0}, we have $g(1)< g(  \alpha^k)$. This contradicts the fact that $  \alpha^k$ minimizes Equation~\eqref{eq:one:d:search}.
Therefore,  $  \alpha^k$ is always positive. 

If $X  \mathbf h^k \neq X \bm \beta^k$, $g(\alpha)$ is strongly convex and thus $ \alpha^k$ is unique.
This ends the proof of this theorem. \hfill $\Box$

For coordinate descent, the condition in Equation~\eqref{eq:cd:non:increasing} always holds, because the objective function 
value keeps decreasing. The selection of an appropriate $\mathbf h^k$ is key to the success of the proposed SRR,
and the following theorem says that if  $\mathbf h^k$ is good enough, the refined search solution  $\mathbf s^k$
is an optimal solution to Equation~\eqref{eq:lasso:problem}.
\begin{theorem}\label{theorem:optimal:beta}
Let $\bm \beta^*$ be an optimal solution to Equation~\eqref{eq:lasso:problem}. 
If 
\begin{equation}\label{eq:refinement:optimal}
   \bm \beta^* -   \mathbf h^k = \gamma ( \bm \beta^k -   \mathbf h^k),
\end{equation}
for some positive $\gamma$, $\mathbf s^k$ achieved by SRR in Equation~\eqref{eq:SGR:scheme}
satisfies $f(\mathbf s^k)=f(\bm \beta^* )$.
\end{theorem}
\noindent \textbf{Proof }
When setting $  \alpha^k=\gamma$, we have $\mathbf s^k=\bm \beta^*$ under the assumption in Equation~\eqref{eq:refinement:optimal}.
Therefore, with the SRR technique, we can obtain a refined solution $\mathbf s^k$
that is an optimal solution to Equation~\eqref{eq:lasso:problem}. \hfill $\Box$

In the following subsections, we discuss two schemes for choosing the history solution $\mathbf h^k$.

\subsection{Successive Ray Refinement Chain}

\begin{figure}[h]
  \centering
	\includegraphics[width=\columnwidth]{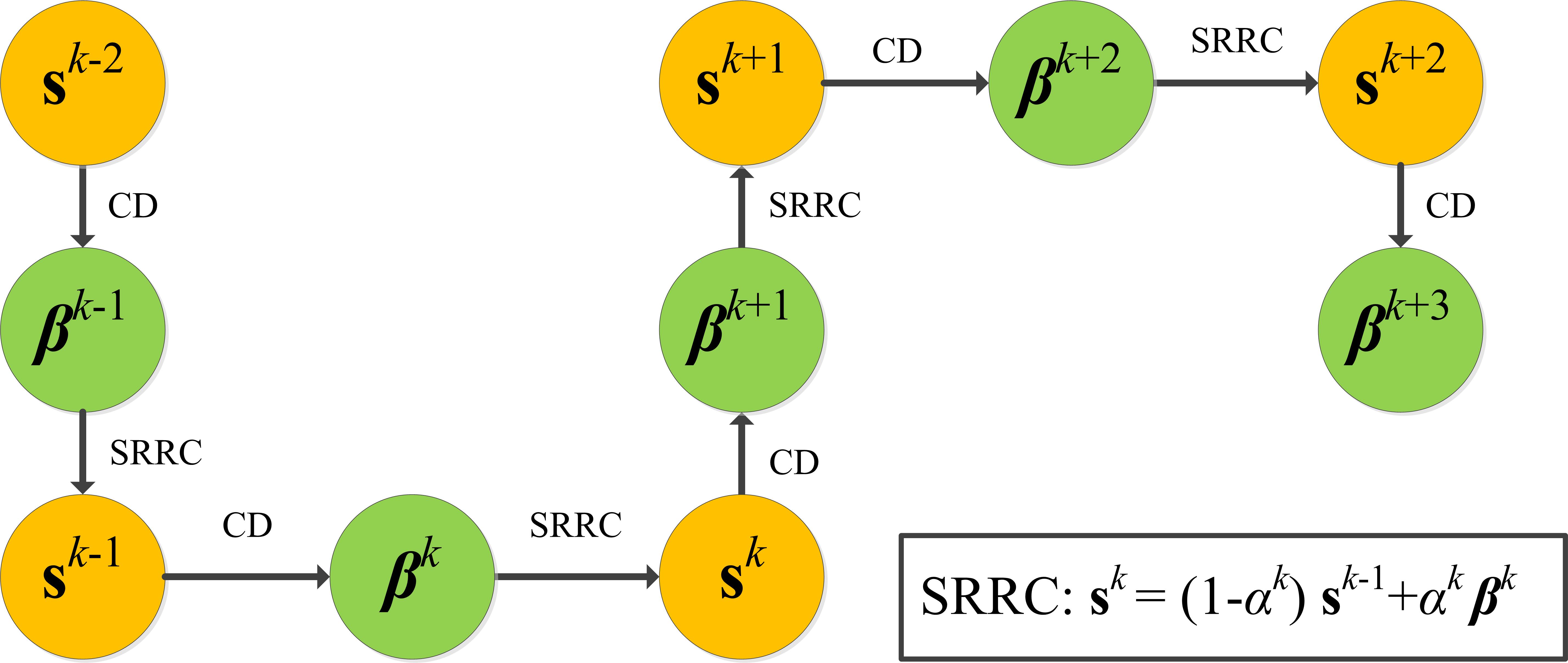}
	\caption{Illustration of the proposed SRRC technique. $\mathbf s^k$ is the refined search point and $\bm \beta^{k+1}$
	is the point that is obtained by applying coordinate descent (CD) based on the refined search point $\mathbf s^k$. In this illustration,
	it is assumed that the optimal refinement factor $\alpha^k$ is larger than 1. When $\alpha^k \in (0,1)$, $\mathbf s^k$ lies 
	between $\mathbf s^{k-1}$ and $\bm \beta^k$.}
	\label{fig:sorc:diagram}
\end{figure}

In the first scheme, we set 
\begin{equation}
    \mathbf h^k = \mathbf s^{k-1}.
\end{equation}
That is, the history point is set to the most recent refined search point.
Figure~\ref{fig:sorc:diagram} demonstrates this scheme.
Since the generated points follow a chain structure,
we call this scheme the Successive Ray Refinement Chain (SRRC).
In SRRC, $\mathbf s^{k-1}$, $\bm \beta^k$, and $\mathbf s^k$ lie on the same line.
In addition, coordinate descent (CD) controls the direction of the chain. In this illustration,
it is assumed that the optimal refinement factor $\alpha^k$ is larger than 1 in each step.
According to Theorem~\ref{theorem:non:negative}, $\alpha^k >0$. When $\alpha^k \in (0,1)$, $\mathbf s^k$ lies 
between $\mathbf s^{k-1}$ and $\bm \beta^k$. When $\alpha^k =1$, $\mathbf s^k$ coincides with $\bm \beta^k$.

In Algorithm~\ref{algorithm:SGR:outer}, we apply the proposed SRRC to coordinate descent for Lasso. 
Compared with the traditional coordinate descent in Algorithm~\ref{algorithm:traditional},
the coordinate update is based on the search point $\mathbf s^{k-1}$ rather than on the previous solution $\bm \beta^{k-1}$. 
When $\alpha^k$ in line 9 of Algorithm~\ref{algorithm:SGR:outer}
is set to 1, Algorithm~\ref{algorithm:SGR:outer} becomes identical to  Algorithm~\ref{algorithm:traditional}.

\begin{algorithm}
\caption{Coordinate Descent plus SRRC (CD+SRRC) for Lasso}        
\label{algorithm:SGR:outer}     
\begin{algorithmic}[1] 
    \REQUIRE $X$, $\mathbf y$, $\lambda$
    \ENSURE $\bm \beta^k$
    \STATE Set $k=0$, $\mathbf s^0 = \mathbf 0$, $\mathbf r_s^0 = \mathbf y$
    \REPEAT
				\STATE Set $k=k+1$, $\mathbf r^k = \mathbf r_s^{k-1}$				
				
        \FOR{$i=1$  to $p$}
				   \STATE Compute $\beta_i^k =   S( s_i^{k-1} + \frac{\mathbf x_i^T \mathbf r^k}{\|\mathbf x_i\|_2^2} , \frac{\lambda} {\|\mathbf x_i\|_2^2})$
				   \STATE Obtain $\mathbf r^k = \mathbf r^k + \mathbf x_i (s_i^{k-1} - \beta_i^k)$
				\ENDFOR

				\IF{convergence criterion not satisfied}
				  \STATE Set $\alpha^k=\arg \min_{\alpha} f( (1-\alpha) \mathbf s^{k-1} + \alpha \bm \beta^k)$
				  \STATE Set $\mathbf s^k = (1-  \alpha^k) \mathbf s^{k-1} +   \alpha^k \bm \beta^k$
				  \STATE Set $\mathbf r_s^k =  (1-  \alpha^k) \mathbf r_s^{k-1} +   \alpha^k \mathbf r^k$
				\ENDIF
    \UNTIL{convergence criterion satisfied}
\end{algorithmic}
\end{algorithm}

\begin{table}
\caption{Illustration of coordinate descent plus SRRC for solving the same problem as in Table~\ref{table:iteration:traditiona:cd}.
Note that the optimal function value is 0.}
\begin{center}
\begin{small}
\begin{tabular}{cccccccc}
  \hline
  $k$ & $\beta_1^k$ & $\beta_2^k$ & $\beta_3^k$ & $\beta_4^k$ & $\beta_5^k$ & $f(\bm \beta^k)$ & $  \alpha^k$ \\
  \hline\hline
1
 & 
0.048912 & 0.034041 & 0.407960 & 0.055687 & 0.160413
 & 
0.052449
\\ \hline
2
 & 
0.058130 & -0.041464 & 0.471828 & 0.024612 & 0.173040
 & 
0.016773
 & 
1.114740
\\ \hline
3
 & 
0.022324 & -0.108065 & 0.459034 & -0.018702 & 0.181180
 & 
0.004209
 & 
1.520601
\\ \hline
4
 & 
-0.000996 & -0.137517 & 0.452455 & -0.033262 & 0.187045
 & 
0.001791
 & 
1.610933
\\ \hline
5
 & 
-0.010602 & -0.144776 & 0.452250 & -0.033032 & 0.190141
 & 
0.001452
 & 
1.114831
\\ \hline
6
 & 
-0.029911 & -0.153851 & 0.453133 & -0.026748 & 0.196740
 & 
0.001091
 & 
2.700667
\\ \hline
7
 & 
-0.047531 & -0.149751 & 0.458275 & -0.006750 & 0.203971
 & 
0.000632
 & 
3.936469
\\ \hline
8
 & 
-0.052347 & -0.148709 & 0.459668 & -0.001392 & 0.205944
 & 
0.000530
 & 
1.398237
\\ \hline
9
 & 
-0.058803 & -0.147299 & 0.461526 & 0.005839 & 0.208586
 & 
0.000407
 & 
2.059226
\\ \hline
10
 & 
-0.064683 & -0.145997 & 0.463220 & 0.012415 & 0.210994
 & 
0.000308
 & 
2.134921
\\ \hline
...
\\ \hline
13
 & 
-0.078615 & -0.142905 & 0.467249 & 0.028058 & 0.216701
 & 
0.000129
 & 
1.100414
\\ \hline
14
 & 
-0.093529 & -0.139593 & 0.471552 & 0.044782 & 0.222808
 & 
0.000023
 & 
9.617055
\\ \hline
15
 & 
-0.094167 & -0.139451 & 0.471743 & 0.045510 & 0.223071
 & 
0.000020
 & 
0.997764
\\ \hline
16
 & 
-0.104249 & -0.137213 & 0.474657 & 0.056823 & 0.227201
 & 
3.3020e-11
 & 
16.530123
\\ \hline
...
\\ \hline
28
 & 
-0.104260 & -0.137210 & 0.474660 & 0.056835 & 0.227205
 & 
6.3207e-15
 & 
1.197748
\\ \hline
29
 & 
-0.104260 & -0.137210 & 0.474660 & 0.056835 & 0.227205
 & 
5.7081e-15
 & 
0.890278
\\ \hline
30
 & 
-0.104260 & -0.137210 & 0.474660 & 0.056835 & 0.227205
 & 
2.0807e-15
 & 
6.945350
\\ \hline
\end{tabular}
\end{small}
\end{center}
\vspace{-0.1in}
\label{table:iteration:SGR:1}
\end{table}

Table~\ref{table:iteration:SGR:1} illustrates Algorithm~\ref{algorithm:SGR:outer} with the same input $X$ and $\mathbf y$ that are
used in Table~\ref{table:iteration:traditiona:cd}.
Comparing Table~\ref{table:iteration:SGR:1} with Table~\ref{table:iteration:traditiona:cd}, we can see that 
the number of iterations can be significantly reduced with the usage of the SRRC technique.
Specifically, to achieve a function value of below $10^{-3}$, the traditional coordinate descent takes 10 iterations, whereas 
the one with the SRRC technique takes 7 iterations; to achieve a function value below $10^{-4}$, 
the traditional coordinate descent takes 29 iterations,
whereas the one with the SRRC technique 14 iterations; and to achieve a function value below $10^{-8}$, 
the traditional coordinate descent takes 103 iterations,
whereas the one with the SRRC technique takes 16 iterations.

As can be seen from Figure~\ref{fig:sorc:diagram}, we generate two sequences: $\{\mathbf s^k\}$ and $\{\bm \beta^k\}$.
At iteration $k$, the SRRC technique is very greedy in that it constructs the search point $\mathbf s^k$ by using the two existing points
$\mathbf s^{k-1}$ and $\bm \beta^k$ to achieve the lowest objective function value. 
If the search point $\mathbf s^{k-1}$ is dense at some iteration number $k$ and $\alpha^k \neq 1$, it can be shown that
$\mathbf s^k$ is also dense. This is not good for Lasso, which usually has a sparse solution.
Interestingly, our empirical simulations show that
Algorithm~\ref{algorithm:SGR:outer} can set
$\alpha^k = 1$ in some iterations, leading to a sparse search point.

\subsection{Successive Ray Refinement Triangle}

\begin{figure}
  \centering
	\includegraphics[width=3in]{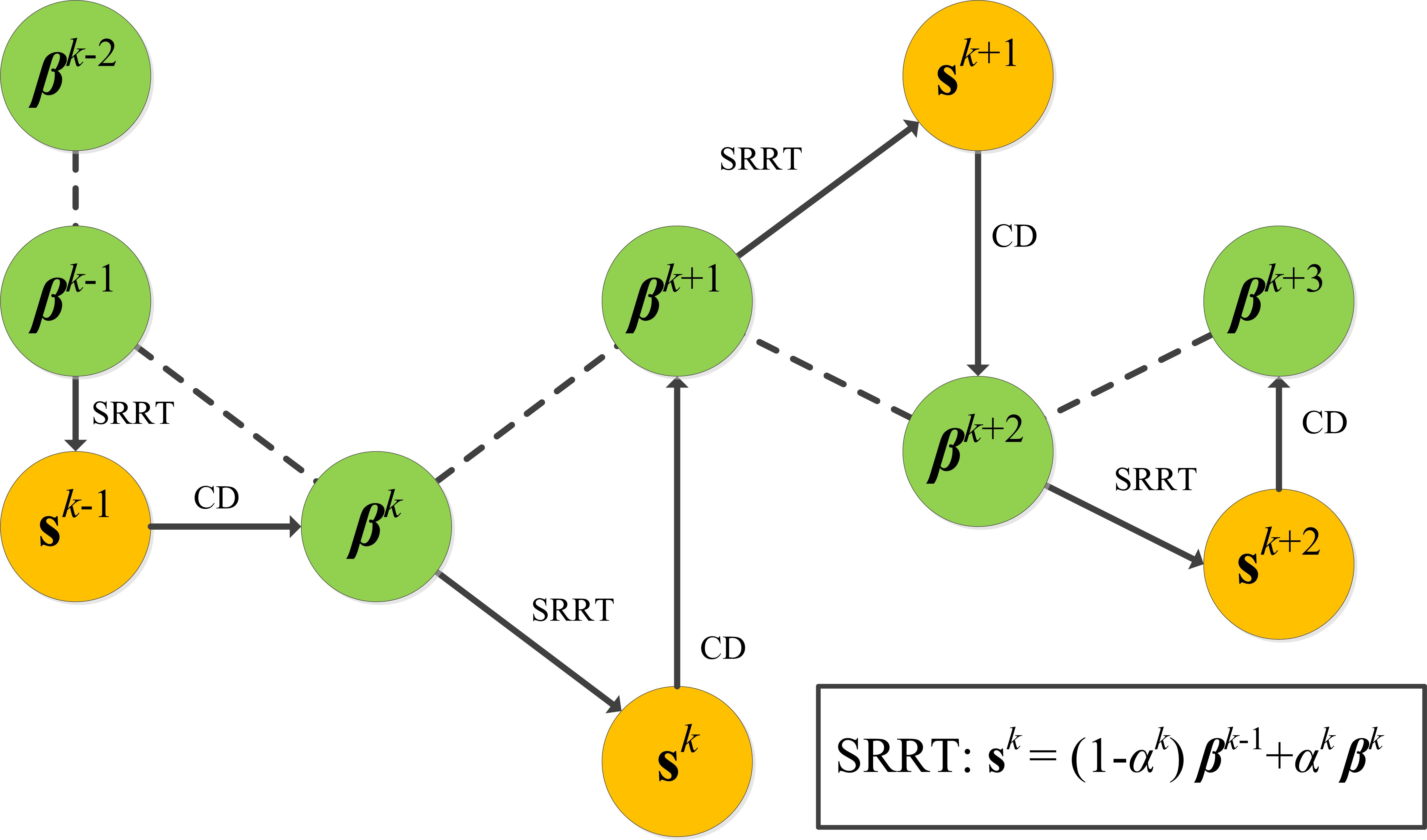}
	\caption{Illustration of the proposed SRRT technique. $\mathbf s^k$ is the refined search point and $\bm \beta^{k+1}$
	is the point obtained by applying coordinate descent (CD) based on the refined search point $\mathbf s^k$. In this illustration,
	it is assumed that the optimal refinement factor $\alpha^k$ is larger than 1. When $\alpha^k \in (0,1)$, $\mathbf s^k$ lies 
	between $\bm \beta^{k-1}$ and $\bm \beta^k$.}
	\label{fig:sort:diagram}
\end{figure}

In the second scheme, we set 
\begin{equation}
    \mathbf h^k = \bm \beta^{k-1}.
\end{equation}
Figure~\ref{fig:sort:diagram} demonstrates this scheme.
Since the generated points follow a triangle structure,
we call this scheme the Successive Ray Refinement Triangle (SRRT).
SRRT is less greedy compared to SRRC because $\bm \beta^{k-1}$ leads to a higher objective function value than $\mathbf s^{k-1}$ leads to. 
However, SRRT can sometimes outperform SRRC in solving Lasso.

Algorithm~\ref{algorithm:orrt} shows the application of the proposed SRRT technique to coordinate descent for Lasso. Similar to Algorithm~\ref{algorithm:SGR:outer},
if $\alpha^k$ in line 9 is set to 1, Algorithm~\ref{algorithm:orrt} reduces to the traditional coordinate descent in Algorithm~\ref{algorithm:traditional}.
Table~\ref{table:iteration:orrt} illustrates Algorithm~\ref{algorithm:orrt}. Similar to SRRC, SRRT greatly reduces the number of iterations used in coordinate descent for Lasso.

\begin{algorithm}   
\caption{Coordinate Descent plus SRRT (CD+SRRT) for Lasso}        
\label{algorithm:orrt}     
\begin{algorithmic}[1] 
    \REQUIRE $X$, $\mathbf y$, $\lambda$
    \ENSURE $\bm \beta^k$
    \STATE Set $k=0$, $\mathbf s^0 = \mathbf 0$, $\mathbf r_s^0 = \mathbf y$
    \REPEAT
				\STATE Set $k=k+1$, $\mathbf r^k = \mathbf r_s^{k-1}$				
				
        \FOR{$i=1$  to $p$}
				   \STATE Compute $\beta_i^k =   S( s_i^{k-1} + \frac{\mathbf x_i^T \mathbf r^k}{\|\mathbf x_i\|_2^2} , \frac{\lambda} {\|\mathbf x_i\|_2^2})$
				   \STATE Obtain $\mathbf r^k = \mathbf r^k + \mathbf x_i (s_i^{k-1} - \beta_i^k)$
				\ENDFOR

				\IF{convergence criterion not satisfied}
				  \STATE Set $\alpha^k= \arg \min_{\alpha} f( (1-\alpha) \bm \beta^{k-1} + \alpha \bm \beta^k)$
				  \STATE Set $\mathbf s^k = (1-  \alpha^k) \bm \beta^{k-1} +   \alpha^k \bm \beta^k$
				  \STATE Set $\mathbf r_s^k =  (1-  \alpha^k) \mathbf r^{k-1} +   \alpha^k \mathbf r^k$
				\ENDIF
    \UNTIL{convergence criterion satisfied}
\end{algorithmic}
\end{algorithm}

\begin{table}
\caption{Illustration of coordinate descent plus SRRT for solving the same problem as in Table~\ref{table:iteration:traditiona:cd}.
Note that the optimal function value is 0.}
\begin{center}
\begin{small}
\begin{tabular}{cccccccc}
  \hline
  $k$ & $\beta_1^k$ & $\beta_2^k$ & $\beta_3^k$ & $\beta_4^k$ & $\beta_5^k$ & $f(\bm \beta^k)$ & $  \alpha^k$ \\
  \hline\hline
1
 & 
0.048912 & 0.034041 & 0.407960 & 0.055687 & 0.160413
 & 
0.052449
\\ \hline
2
 & 
0.058130 & -0.041464 & 0.471828 & 0.024612 & 0.173040
 & 
0.016773
 & 
1.114740
\\ \hline
3
 & 
0.032838 & -0.089244 & 0.463272 & -0.006319 & 0.178907
 & 
0.006746
 & 
1.077199
\\ \hline
4
 & 
-0.010078 & -0.154209 & 0.449373 & -0.043957 & 0.189153
 & 
0.001610
 & 
2.336008
\\ \hline
5
 & 
-0.015176 & -0.152741 & 0.450482 & -0.038189 & 0.191151
 & 
0.001435
 & 
0.960038
\\ \hline
6
 & 
-0.087427 & -0.134480 & 0.471220 & 0.044793 & 0.220728
 & 
0.000061
 & 
15.373834
\\ \hline
7
 & 
-0.098214 & -0.134199 & 0.474324 & 0.054977 & 0.225125
 & 
0.000019
 & 
1.138984
\\ \hline
8
 & 
-0.104044 & -0.135210 & 0.475348 & 0.058970 & 0.227325
 & 
0.000005
 & 
1.492143
\\ \hline
9
 & 
-0.106491 & -0.136042 & 0.475553 & 0.060121 & 0.228188
 & 
0.000002
 & 
1.414778
\\ \hline
10
 & 
-0.106739 & -0.136361 & 0.475482 & 0.059962 & 0.228250
 & 
0.000002
 & 
1.141313
\\ \hline
...
\\ \hline
16
 & 
-0.104212 & -0.137315 & 0.474613 & 0.056673 & 0.227176
 & 
1.2569e-08
 & 
1.246072
\\ \hline
17
 & 
-0.104115 & -0.137272 & 0.474607 & 0.056637 & 0.227143
 & 
6.9462e-09
 & 
1.481390
\\ \hline
18
 & 
-0.104106 & -0.137256 & 0.474611 & 0.056648 & 0.227141
 & 
5.5802e-09
 & 
1.169237
\\ \hline
...
\\ \hline
28
 & 
-0.104268 & -0.137209 & 0.474662 & 0.056843 & 0.227209
 & 
1.1516e-11
 & 
1.808904
\\ \hline
29
 & 
-0.104264 & -0.137210 & 0.474660 & 0.056839 & 0.227207
 & 
3.9831e-12
 & 
4.426169
\\ \hline
30
 & 
-0.104262 & -0.137210 & 0.474660 & 0.056836 & 0.227206
 & 
1.1488e-12
 & 
1.651176
\\ \hline
\end{tabular}
\end{small}
\end{center}
\vspace{-0.1in}
\label{table:iteration:orrt}
\end{table}

\subsection{Convergence of CD plus SRR}

In this subsection, we show that both the combination of CD and SRRC (CD+SRRC)  and the combination of CD and SRRT (CD+SRRT) are guaranteed to converge.

\begin{theorem}\label{theorem:convergence}
For the sequence $\mathbf s^0, \bm \beta^1 ,\mathbf s^1, \bm \beta^2, \mathbf s^2, \bm \beta^3 ,\ldots$ generated by CD+SRRC and CD+SRRT,
the objective function value is monotonically decreasing until convergence; that is, 
\begin{equation}
   f(\mathbf s^{k-1}) \ge f(\bm \beta^k) \ge f(\mathbf s^k) \ge f(\bm \beta^{k+1}).
\end{equation}
In addition, if $f(\mathbf s^{k-1}) = f(\bm \beta^k)$, we have $\mathbf s^{k-1}=\bm \beta^k$ and $\bm \beta^k$ is an optimal solution; that is,
\begin{equation}
    f(\bm \beta^k) = \min_{\bm \beta} f(\bm \beta).
\end{equation}
Therefore, we have
\begin{equation}\label{eq:convergence:mononotone}
   \lim_{k \rightarrow \infty} f(\bm \beta^k) = \min_{\bm \beta} f(\bm \beta).
\end{equation}
\end{theorem}

\noindent \textbf{Proof } $\bm \beta^k$ is computed by applying coordinate descent based on $\mathbf s^{k-1}$; that is,
\begin{equation}
   \beta_i^k= \arg \min_{\beta} f([ \beta_1^k,\ldots,\beta_{i-1}^k,\beta,s_{i+1}^{k-1},\ldots,s_{p}^{k-1}   ]^T),
\end{equation}
or equivalently
\begin{equation} \label{eq:soft:prove}
\beta_i^k = \frac{ S(\mathbf x_i^T \mathbf y - \sum_{j<i} \mathbf x_i^T \mathbf x_j \beta_j^k - \sum_{j>i} \mathbf x_i^T \mathbf x_j s_j^{k-1} , \lambda) }{\|\mathbf x_i\|_2^2}.
\end{equation}
Therefore, we have
\begin{equation}\label{eq:f:b:s}
\begin{aligned}
    & f([ \beta_1^k,\ldots,\beta_{i-1}^k,\beta_i^k,s_{i+1}^{k-1},\ldots,s_{p}^{k-1}   ]^T) \\
	 & \le f([ \beta_1^k,\ldots,\beta_{i-1}^k,s_i^{k-1},s_{i+1}^{k-1},\ldots,s_{p}^{k-1}   ]^T).
\end{aligned}
\end{equation}
for all $i$. 

Since $\|\mathbf x_i\|_2 \neq 0$, $f([ \beta_1^k,\ldots,\beta_{i-1}^k,\beta,s_{i+1}^{k-1},\ldots,s_{p}^{k-1}   ]^T)$ is strongly convex in $\beta$.
As a result, if the equality in Equation~\eqref{eq:f:b:s} holds, we have $\beta_i^k=s_i^{k-1}$.
Recursively applying Equation~\eqref{eq:f:b:s}, we have the following two facts: $f(\bm \beta^k) \le f(\mathbf s^{k-1})$ and if $f(\bm \beta^k) = f(\mathbf s^{k-1})$, then $\mathbf s^{k-1}=\bm \beta^k$.

If $\mathbf s^{k-1}=\bm \beta^k$, it follows from Equation~\eqref{eq:soft:prove} that
\begin{equation} 
\begin{aligned}
\|\mathbf x_i\|_2^2 \beta_i^k & =  S(\mathbf x_i^T \mathbf y - \sum_{j \neq i} \mathbf x_i^T \mathbf x_j \beta_j^k   , \lambda) \\
                              & =  S(\mathbf x_i^T \mathbf y - \mathbf x_i^T  X \bm \beta^k + \|\mathbf x_i\|_2^2 \beta_i^k  , \lambda),
\end{aligned}
\end{equation}
which leads to
\begin{equation}\label{eq:beta:k:optimal}
    \mathbf x_i^T \mathbf y - \mathbf x_i^T  X \bm \beta^k  \in \mbox{SGN}(\beta_i^k),
\end{equation}
where
\begin{equation}
\mbox{SGN}(t)=
\left\{
\begin{array}{cc}
  \{ 1  \},  & t >0 \\
	\{ -1 \},  & t <0 \\
	\left[-1,1\right], & t =0.  \\
\end{array}
\right.
\end{equation}
Since $\bm \beta^*$ is an optimal solution to Equation~\eqref{eq:lasso:problem} if and only if
\begin{equation}
    \mathbf x_i^T \mathbf y - \mathbf x_i^T  X \bm \beta^*  \in \mbox{SGN}(\beta_i^*), \forall i,
\end{equation}
it follows from Equation~\eqref{eq:beta:k:optimal} that $\bm \beta^k$ is an optimal solution to Equation~\eqref{eq:lasso:problem}.

The relationship $f(\bm \beta^k) \ge f(\mathbf s^k)$ is guaranteed by the univariate optimization problem in Equation~\eqref{eq:one:d:search}.
Therefore, the sequence $\{f(\bm \beta^k)\}$ is decreasing. Meanwhile, the squence $\{f(\bm \beta^k)\}$ has a lower bound $\min_{\bm \beta} f(\bm \beta)$.
According to the well-known monotone convergence theorem, we have Equation~\eqref{eq:convergence:mononotone}.

This completes the proof of this theorem. \hfill $\Box$

%% file: alphaCompute.tex
In this section, we discuss how to efficiently compute the refinement factor $\alpha^k$
in Equation~\eqref{eq:one:d:search}. The function $g(\alpha)$ can be written as:
\begin{equation}
\begin{aligned}
   g(\alpha) & = \frac{1}{2} \left\| X ((1-\alpha) \mathbf h^k + \alpha \bm \beta^k) - \mathbf y \right\|_2^2 + \lambda \|(1-\alpha)   \mathbf h^k + \alpha \bm \beta^k\|_1 \\
	           & = \frac{1}{2} \left\|  \mathbf r^k_h- \alpha (  \mathbf r^k_h- \mathbf r^k)  \right\|_2^2 + \lambda \| \mathbf h^k-\alpha ( \mathbf h^k - \bm \beta^k)\|_1,
\end{aligned}
\end{equation}
where $\mathbf r^k_h = \mathbf y - X  \mathbf h^k$ and $\mathbf r^k=\mathbf y - X   \bm \beta^k$
are the residuals that correspond to $ \mathbf h^k$ and $\bm \beta^k$, respectively. Note that 1) $\mathbf r^k_h = \mathbf r^{k-1}_s$ for SRRC and
$\mathbf r^k_h = \mathbf r^{k-1}$ for SRRT, and 2) both $\mathbf r^k_h$ and $\mathbf r^k$ have been obtained before line 8 of Algorithm~\ref{algorithm:SGR:outer}
and Algorithm~\ref{algorithm:orrt}. Before the convergence, we have $\mathbf r^k_h \neq \mathbf r^k$. Therefore, $g(\alpha)$ is strongly convex in $\alpha$, and
$\alpha^k$, the minimizer to Equation~\eqref{eq:one:d:search}, is unique.

When $\lambda=0$, Equation~\eqref{eq:one:d:search} has a nice closed form solution,
\begin{equation}
     \alpha^k = \frac{ \langle \mathbf r^k_h  , \mathbf r^k_h - \mathbf r^k  \rangle }{\|  \mathbf r^k_h - \mathbf r^k \|_2^2}.
\end{equation}

\begin{figure}
\centering
	 \includegraphics[width=0.7\columnwidth]{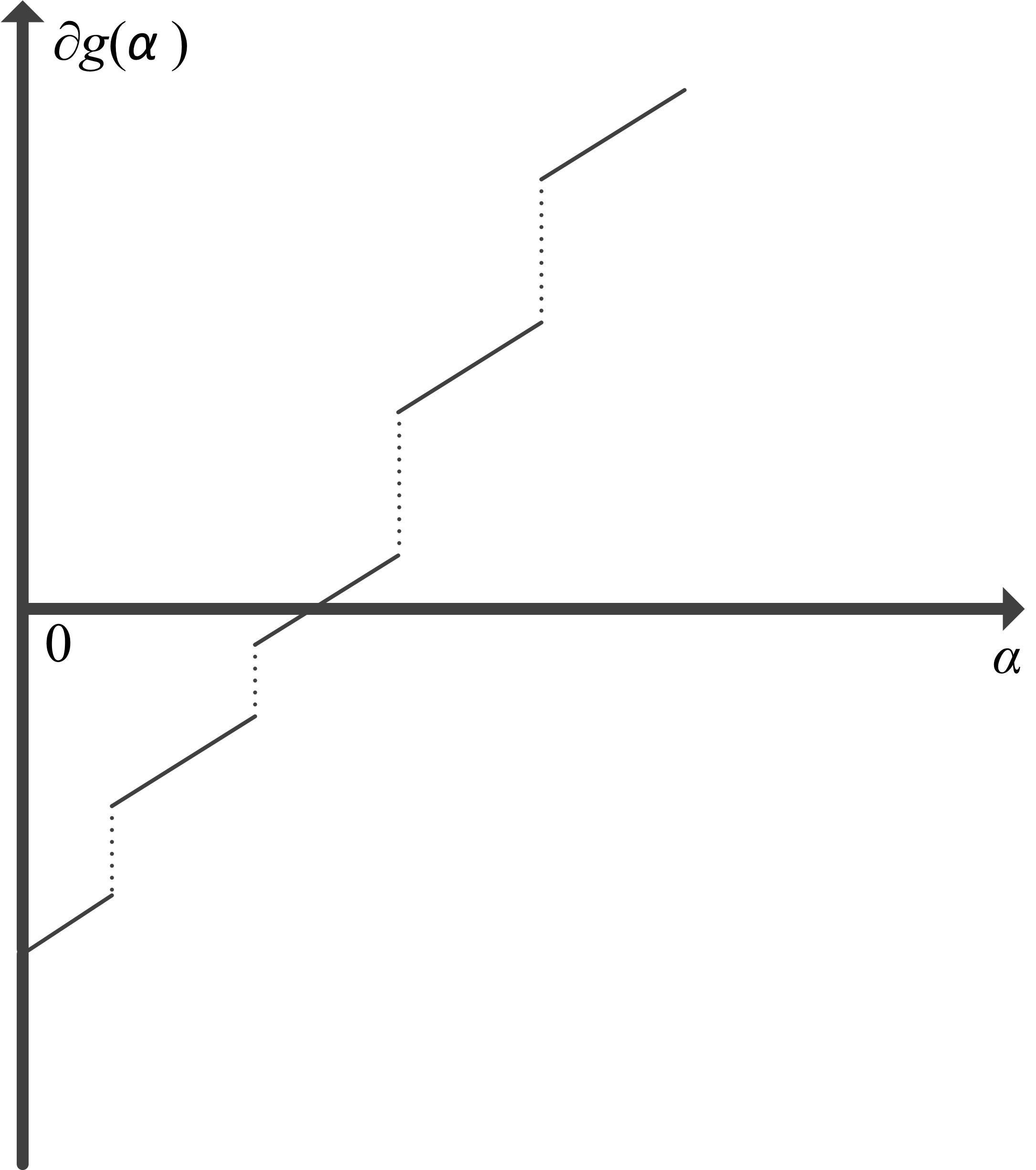}
	\caption{Illustration of $\partial g(\alpha)$. When $\lambda >0$, it is a non-continuous piecewise linear monotonically increasing function. The intersection between $\partial  g(\alpha)$
	and the horizontal axis gives $\tilde \alpha^k$, the solution to Equation~\eqref{eq:one:d:search}.}
	\label{fig:sfg:computation}
\end{figure}

Next, we discuss the case $\lambda >0$. The subgradient of $ g(\alpha)$ with regard to $\alpha$ can be computed as
\begin{equation}
\begin{aligned}
\partial  g(\alpha) & =   \alpha \| \mathbf r^k_h - \mathbf r^k\|_2^2 -  \langle  \mathbf r^k_h , \mathbf r^k_h  - \mathbf r^k \rangle \\
                    &   + \lambda \sum_{i=1}^p (\beta^k_i - h_i) \mbox{SGN} (h_i-\alpha ( h_i -  \beta^k_i)).
\end{aligned}
\end{equation}
Compute $\alpha^k$ is a root-finding problem.
According to Theorem~\ref{theorem:non:negative}, we have $\alpha^k >0$. Next, we consider only $\alpha>0$ for $\partial  g(\alpha)$.
We consider the following three cases:
\begin{itemize}
  \item[1.] If $h_i =0$, we have $$(\beta^k_i - h_i) \mbox{SGN} (h_i-\alpha ( h_i -  \beta^k_i)) = \{ |\beta^k_i| \}.$$
  \item[2.] If $h_i (\beta_i^k - h_i) >0$, we have $$(\beta^k_i - h_i) \mbox{SGN} (h_i-\alpha ( h_i -  \beta^k_i)) = \{ |\beta^k_i - h_i| \}.$$
  \item[3.] If $h_i (\beta_i^k - h_i) <0$, we let 	
	\begin{equation}
	   w_i = \frac{h_i}{ h_i - \beta^k_i },
	\end{equation}
	 and we have
	   \begin{equation}
		  \begin{aligned}
		     & (\beta^k_i - h_i) \mbox{SGN} (h_i-\alpha ( h_i -  \beta^k_i)) = \\
				& \left\{ 
				 \begin{array}{ll}
\{ -|\beta^k_i - h_i| \}      & \alpha \in (0, w_i) \\
 \{ |\beta^k_i - h_i| \}      & \alpha \in (w_i, +\infty) \\
|\beta^k_i - h_i| \{ [-1,1]\} & \alpha =w_i. \\
         \end{array} \right.
		 \end{aligned}
		 \end{equation}
\end{itemize}
For the first two cases, the set $\mbox{SGN} (h_i-\alpha ( h_i -  \beta^k_i))$ is deterministic. For the third case, 
$\mbox{SGN} (h_i-\alpha ( h_i -  \beta^k_i))$ is deterministic when  $\alpha \neq w_i$.
Define
\begin{equation}
  \Omega(\mathbf h^k, \bm \beta^k) = \{i: h_i (\beta_i^k - h_i) <0 \}.
\end{equation}
Figure~\ref{fig:sfg:computation} illustrates the function $\partial  g(\alpha), \alpha >0$. 
It can be observed that $\partial  g(\alpha)$ is a piecewise linear function.
If $\Omega(\mathbf h^k, \bm \beta^k)$ is empty, $\partial  g(\alpha)$ is continuous; otherwise, $\partial  g(\alpha)$
is not continuous at $\alpha=w_i, i \in \Omega(\mathbf h^k, \bm \beta^k)$.

\subsection{An Algorithm Based on Sorting}

To compute the refinement factor, one approach is to sort $w_i$ as follows: 

First, we sort $w_i, i \in  \Omega(\mathbf h^k, \bm \beta^k)$, and assume $w_{i_0} \le w_{i_1} \le \ldots \le w_{i_{|\Omega(\mathbf h^k, \bm \beta^k)|}}$.

Second, for $j=1, 2, \ldots, |\Omega(\mathbf h^k, \bm \beta^k)|$, we evaluate $\partial  g(\alpha)$ at $\alpha = w_{i_j}$ with the following three cases:
\begin{itemize}
   \item[1.]  If $ 0 \in \partial  g(w_{i_j})$, we have $\alpha^k = w_{i_j}$ and terminate the search.
	 \item[2.]  If an element in $\partial  g(w_{i_j})$ is positive, $\alpha^k$ lies in the piecewise line starting $\alpha=w_{i_{j-1}}$ and ending $\alpha=w_{i_j}$, and it can be analytically computed.
	 \item[3.]  If all elements in $\partial  g(w_{i_j})$ are negative, we set $j=j+1$ and continue the search.
\end{itemize}
Finally, if all elements in $\partial  g(w_{i_j})$ are negative when $j=|\Omega(\mathbf h^k, \bm \beta^k)|$, 
$\alpha^k$ lies on the piecewise line that starts at $\alpha=w_{i_j}$. Thus,
$\tilde \alpha^k$ can be analytically computed.

With a careful implementation, the naive approach can be completed in $O(p + m\log(m))$, where $m = |\Omega(\mathbf h^k, \bm \beta^k)| $.
In Lasso, the solution is usually sparse, and thus $m$ is much smaller than $p$, the number of variables.

\subsection{An Algorithm Based on Bisection}

A second approach is to make use of the improved bisection proposed in~\cite{liu_j:09}.
The idea is to 1) determine an initial guess of the interval $[\alpha_1, \alpha_2]$ to which the root belongs,
where all elements in $\partial  g(\alpha_1)$
are negative and all elements in $\partial  g(\alpha_2)$ are positive, 2) evaluate
$\partial g(\alpha)$ at $\alpha= \frac{\alpha_1+\alpha_2}{2}$ and update 
the interval to $[\alpha_1, \alpha)$ if all the elements in $\partial g(\alpha)$ are positive
or to $[\alpha, \alpha_2)$ if all the elements in $\partial  g(\alpha)$
are negative, 3) set the value of $\alpha$ to the largest value of $w_i$ that satisfy $w_i < \alpha$ if  all the elements in $\partial  g(\alpha)$ are positive
or to the smallest value of $w_i$ that satisfy $w_i > \alpha$ if  all the elements in $\partial  g(\alpha)$ are negative,
and 4) repeat 2) and 3) until finding the root of $\partial  g(\alpha)$.
With a similar implementation as in~\cite{liu_j:09}, the improved bisection approach has a time complexity of $O(p)$.


%% file: justification.tex
Let 
\begin{equation}\label{eq:LDU:XTX}
A= X X^T = L + D + U,
\end{equation}
where $D$ is $A$'s diagonal part, $L$ is $A$'s strictly lower triangular part, and $U$ is $A$'s strictly upper triangular part.
It is easy to see that
\begin{equation}
L_{ij}=
	\left\{
\begin{array}{cc} 
 \mathbf x_i^T \mathbf x_j  &  i<j \\
 0 &  i  \ge j ,\\
\end{array}		\right.
\end{equation}
\begin{equation}
D_{ij}=
	\left\{
\begin{array}{cc} 
 \mathbf x_i^T \mathbf x_i  &  i=j \\
 0 &  i \neq j ,\\
\end{array}		\right.
\end{equation}
\begin{equation}
U_{ij}=
	\left\{
\begin{array}{cc} 
 \mathbf x_i^T \mathbf x_j  &  i>j \\
 0 &  i  \le j. \\
\end{array}		\right.
\end{equation}

We can rewrite Equation~\eqref{eq:beta:i:kplus1} as
\begin{equation}\label{eq:beta:i:kplus1:reformulation}
D_{ii} \beta_i^k =   S(\mathbf x_i^T \mathbf y - L_{i:} \bm \beta^k - U_{i:}\bm \beta^{k-1}, \lambda),
\end{equation}
where $L_{i:}$ and $U_{i:}$ denote the $i$th row of $L$ and $U$, respectively.
Therefore, we can write coordinate descent iteration as:
\begin{equation}\label{eq:beta:i:kplus1:reformulation:2}
D \bm \beta^k =   S(X^T \mathbf y - L \bm \beta^k - U \bm \beta^{k-1}, \lambda).
\end{equation}

When $\lambda=0$, Equation~\eqref{eq:beta:i:kplus1:reformulation:2} becomes
\begin{equation}
  (L +D) \bm \beta^{k+1} =    X^T \mathbf y  - U \bm \beta^k,
\end{equation}
which is the Gauss-Seidel method for solving 
\begin{equation}\label{eq:gradient:zero}
  X^TX \bm \beta = (L+D+U) \bm \beta = X^T \mathbf y.
\end{equation}
Equation~\eqref{eq:gradient:zero} is also the optimality condition for Equation~\eqref{eq:lasso:problem} when $\lambda=0$.
Our next discussion is for the case $\lambda=0$ because it is easy to write the linear systems for the iterations.

Denote 
\begin{equation}\label{eq:G:def}
   G = - (L +D)^{-1} U.
\end{equation}
Let $G$ have the following eigendecomposition:
\begin{equation}
   G = P \Delta P^{-1},
\end{equation}
where $\Delta =\mbox{diag}(\delta_1, \delta_2, \ldots, \delta_p)$ is a diagonal matrix consisting of its eigenvalues.
\begin{lemma}
  The magnitudes of the eigenvalues of $G$ are all less than or equal to 1; that is, 
\begin{equation}
  |\delta_i| \le 1, \forall i.
\end{equation}
\end{lemma}
\noindent \textbf{Proof } Let 
\begin{equation}
     G \mathbf z = \sigma \mathbf z,
\end{equation}
where $\sigma$ is an eigenvalue of $G$ with the corresponding eigenvector being $\mathbf z$. Note that $\sigma$ and
the entries in $\mathbf z$ can be complex.
Using Equation~\eqref{eq:LDU:XTX} and Equation~\eqref{eq:G:def}, 
we have
\begin{equation}
     (L+D - X^TX) \mathbf z  = - U \mathbf z = (L+D) \sigma \mathbf z.
\end{equation}
which leads to
\begin{equation}
     (L+D) (1- \sigma) \mathbf z  = X^TX \mathbf z.
\end{equation}
If $\sigma =1$, the corresponding eigenvector $\mathbf z$ is in the null space of $ X^TX$.
If $\sigma \neq 1$, we have
\begin{equation}\label{eq:proof:less:than:1:1}
     (L+D)  \mathbf z  = \frac{1}{(1- \sigma)} X^TX \mathbf z.
\end{equation}
Premultiplying Equation~\eqref{eq:proof:less:than:1:1} by $\mathbf z^H$, the conjugate transpose of $\mathbf z$,
we have
\begin{equation}\label{eq:proof:less:than:1:2}
     \mathbf z^H (L+D)  \mathbf z  = \frac{1}{(1- \sigma)} \mathbf z^H X^TX \mathbf z.
\end{equation}
Taking the conjugate transpose of Equation~\eqref{eq:proof:less:than:1:2}, we have
\begin{equation}\label{eq:proof:less:than:1:3}
     \mathbf z^H (U+D)  \mathbf z  = \frac{1}{(1- \bar{\sigma})} \mathbf z^H X^TX \mathbf z,
\end{equation}
where $\bar {\sigma}$ denotes the conjugate of $\sigma$.
Adding Equation~\eqref{eq:proof:less:than:1:2} and Equation~\eqref{eq:proof:less:than:1:3} and subtracting $\mathbf z^H X^TX \mathbf z$, 
we have
\begin{equation}\label{eq:proof:less:than:1:4}
     \mathbf z^H D  \mathbf z  = \left(\frac{1}{(1- \bar{\sigma})} + \frac{1}{(1-  \sigma)} -1 \right )\mathbf z^H X^TX \mathbf z.
\end{equation}
Since $\mathbf z^H D  \mathbf z >0$ and $\mathbf z^H X^TX \mathbf z \ge 0$, we have
\begin{equation}
   0 < \frac{1}{(1- \bar{\sigma})} + \frac{1}{(1-  \sigma)} -1 = \frac{1- |\sigma|^2}{|1-\sigma|^2}.
\end{equation}
Therefore, we have $1- |\sigma|^2 >0$ or equivalently $|\sigma| < 1$.

This ends the proof of this lemma. \hfill $\Box$

\subsection{An Eigenvalue Analysis on CD+SRRC}\label{ss:srrc:analysis}

For CD+SRRC in Algorithm~\ref{algorithm:SGR:outer}, when $\lambda =0$ we have
\begin{equation}
   \bm \beta^k = (L+D)^{-1}[X^T \mathbf y - U \mathbf s^{k-1}]
\end{equation}
\begin{equation}\label{eq:sgr:iteration}
   \mathbf s^k =  (1-  \alpha^k)\mathbf s^{k-1} +  \alpha^k \bm \beta^k.
\end{equation}
It can be shown that
\begin{equation}
   \mathbf s^k - \mathbf s^{k-1}=   \alpha^k \left [ - (L +D)^{-1} U + \frac{1-  \alpha^{k-1}}{ \alpha^{k-1}} I \right] (\mathbf s^{k-1} - \mathbf s^{k-2}),
\end{equation}
\begin{equation}
   \bm \beta^k - \bm \beta^{k-1}=  - (L +D)^{-1} U  (\mathbf s^{k-1} - \mathbf s^{k-2}).
\end{equation}

When $k\ge 2$, we denote
\begin{equation}
   A^k =   \alpha^k \left [ - (L +D)^{-1} U + \frac{1-  \alpha^{k-1}}{ \alpha^{k-1}} I \right].
\end{equation}
It can be shown that
\begin{equation}
   A^k =    P   \Sigma^k P^{-1},
\end{equation}
where $ \Sigma^k=\mbox{diag}(\sigma_1^k, \sigma_2^k, \ldots, \sigma_p^k)$ is a diagonal matrix and 
\begin{equation}
   \sigma_i^k =   \alpha^k (\delta_i + \frac{1-  \alpha^{k-1}}{  \alpha^{k-1}}).
\end{equation}
Therefore, we have
\begin{equation}
    \mathbf s^k -  \mathbf s^{k-1}=\left(\Pi_{i=2}^k A^k \right)   ( \mathbf s^1 -  \mathbf s^0) = P (\Pi_{i=2}^k   \Sigma_k) P^{-1} ( \mathbf s^1 -  \mathbf s^0),
\end{equation}

For discussion convenience, we let $\Sigma^1 = \Delta$ and
\begin{equation}
   T^k=\mbox{diag}(t_1^k, t_2^k, \ldots, t_p^k) = \Pi_{i=1}^k   \Sigma^k,
\end{equation}
where
\begin{equation}
   t_i^k =   \Pi_{i=1}^k   \sigma_i^k.
\end{equation}
We have
\begin{equation}
   \bm \beta^k - \bm \beta^{k-1}=  P T^{k-1} P^{-1} ( \mathbf s^1 -  \mathbf s^0).
\end{equation}

For the traditional coordinate descent in Algorithm~\ref{algorithm:traditional}, $\alpha^k =1, \forall k$.
For the proposed CD+SRRC in Algorithm~\ref{algorithm:SGR:outer}, $\alpha^k$ optimizes Equation~\eqref{eq:one:d:search}.

For the data used in Table~\ref{table:iteration:traditiona:cd}, 
the eigenvalues of $ - (L +D)^{-1} U$ are 
$$\delta_1=0, \delta_2=0.00219338, \delta_3=0.12412229,$$
$$  \delta_4=0.62606165,  \delta_5= 0.93956707.$$
For the traditional coordinate descent in Algorithm~\ref{algorithm:traditional}, when $k=30$,
we have
$$t_1^{29} =0, t_2^{29}<0.000001, t_3^{29} <0.000001,$$
$$ t_4^{29}=0.000002, t_5^{29}=0.164023.$$
For the coordinate descent with SRRC in Algorithm~\ref{algorithm:SGR:outer}, we have
$$t_1^{29} =0, t_2^{29}<0.000001, t_3^{29} <0.000001,$$
$$ t_4^{29}<0.000001, t_5^{29}=0.000008.$$
This explains why the proposed SRRC can greatly accelerate the convergence of the coordinate descent method.

\subsection{An Eigenvalue Analysis on CD+SRRT}

For coordinate descent with SRRT in Algorithm~\ref{algorithm:orrt},  when $\lambda =0$ we have
\begin{equation}
   \bm \beta^k = (L+D)^{-1}[X^T \mathbf y - U \mathbf s^{k-1}]
\end{equation}
\begin{equation} 
   \mathbf s^k =  (1-  \alpha^k)\bm \beta^{k-1} +  \alpha^k \bm \beta^k.
\end{equation}
When $k \ge 3$, it can be shown that
\begin{equation}\label{eq:recursion:srrt}
   \bm \beta^k - \bm \beta^{k-1} = G [ (1- \alpha^{k-2}) (\bm \beta^{k-2} - \bm \beta^{k-3}) + \alpha^{k-1}  (\bm \beta^{k-1} - \bm \beta^{k-2})].
\end{equation}
When $k = 2$, we have
\begin{equation}
   \bm \beta^2 - \bm \beta^1 = \alpha^1 G (\bm \beta^1 - \bm \beta^0).
\end{equation}
Using the recursion in Equation~\eqref{eq:recursion:srrt}, we can get
\begin{equation}
   \bm \beta^3 - \bm \beta^2 = [(1 -\alpha^1) G + \alpha^1 G \alpha^2 G ] (\bm \beta^1 - \bm \beta^0).
\end{equation}
\begin{equation}
\begin{aligned}
   \bm \beta^4 - \bm \beta^3 = & [\alpha^1 G (1 -\alpha^2) G +  (1 -\alpha^1) G \alpha^3 G \\
	   & +  \alpha^1 G \alpha^2 G \alpha^3 G]
	   (\bm \beta^1 - \bm \beta^0).
\end{aligned}
\end{equation}
Generally speaking, we can write
\begin{equation} 
   \bm \beta^k - \bm \beta^{k-1} = P T^{k-1} P^{-1} (\bm \beta^1 - \bm \beta^0),
\end{equation}
where $T^k=\mbox{diag}(t_1^k, t_2^k, \ldots, t_p^k)$ is a diagonal matrix.
For $t_i^k$, it is a polynomial function of $\delta_i$; that is, $t_i^k = \phi_k(\delta_i)$, where
\begin{equation}
    \phi_k(t) = \underbrace{t \times \ldots \times t }_{\lceil k/2 \rceil} \sum_{i=0}^{k  - \lceil k/2 \rceil} c_i \times \underbrace{t \times \ldots \times t }_{i} ),
\end{equation}
and $c_0, c_1, \ldots, c_{k  - \lceil k/2 \rceil}$ are dependent on $\alpha^1, \alpha^2, \ldots, \alpha^{k-1}$.
When $k=2$, we have 
\begin{equation}
   c_0= \alpha^1.
\end{equation}
When $k=3$, we have 
\begin{equation}
\begin{array}{lll}
   c_0 &= & 1- \alpha^1, \\
	 c_1 &= & \alpha^1 \alpha^2.\\
\end{array}
\end{equation}
When $k=4$, we have 
\begin{equation}
\begin{array}{lll}
   c_0 &= &  \alpha^1 (1-\alpha^2) + \alpha^3(1-\alpha^1), \\
	 c_1 &= & \alpha^1 \alpha^2 \alpha^3.\\
\end{array}
\end{equation}
When $k=5$, we have 
\begin{equation}
\begin{array}{lll}
   c_0 &=& (1-\alpha^2) (1-\alpha^3), \\
	 c_1 &=&  \alpha^1 (1-\alpha^2) \alpha^4 + \alpha^3(1-\alpha^1) \alpha^4 +  \alpha^1  \alpha^2 (1-\alpha^3),\\
	 c_2 &=& \alpha^1 \alpha^2 \alpha^3 \alpha^4.
\end{array}
\end{equation}
For the coordinate descent with SRRT in Algorithm~\ref{algorithm:orrt}, we have
$$t_1^{29} =0, t_2^{29}<0.000001, t_3^{29} <0.000001, $$
$$t_4^{29}=0.000002, t_5^{29}=0.000393,$$
which are smaller than the ones in the traditional coordinate descent shown in Section~\ref{ss:srrc:analysis}.

%% file: relatedWork.tex
In this section, we compare our proposed SRR with successive over-relaxation~\cite{Yong_D:1950}
and the accelerated gradient descent method~\cite{Nesterov:2004}.

\subsection{Relationship between SRRC and\\ Successive Over-Relaxation}

Successive over-relaxation (SOR) is a classical approach for accelerating the Gauss-Seidel approach. Our discussion in this section considers only $\lambda=0$, because SOR targets the acceleration of the Gauss-Seidel approach.

From Equation~\eqref{eq:gradient:zero}, we have
\begin{equation}\label{eq:gradient:zero:sor}
 (wL+D) \bm \beta = w X^T \mathbf y  - w U \bm \beta  - (w-1) D  \bm \beta, \forall w >0 .
\end{equation}
The iteration used in successive over-relaxation is:
\begin{equation}\label{eq:sor:formulation}
   \bm \beta^k = (wL+D)^{-1} [w X^Ty - [wU + (w-1) D ] \bm \beta^{k-1},
\end{equation}
which can be obtained by plugging $\bm \beta^k $ and $\bm \beta^{k-1}$ into Equation~\eqref{eq:gradient:zero:sor}.
Equation~\eqref{eq:sor:formulation} can be rewritten as:
\begin{equation}\label{eq:sor:formulation:2}
   \bm \beta^k =\bm \beta^{k-1} - w (wL+D)^{-1} [X^TX \bm \beta^{k-1}  - X^T \mathbf y].
\end{equation}

For the proposed CD+SRRC in Algorithm~\ref{algorithm:SGR:outer}, when $\lambda =0$ we have
\begin{equation}\label{eq:srrc:iteration}
\begin{aligned}
  \mathbf s^k & =  (1-  \alpha^k)\mathbf s^{k-1} +   \alpha^k (L +D)^{-1} \left[  X^T \mathbf y  -  U \mathbf s^{k-1} \right] \\
	            & =  \mathbf s^{k-1} -   \alpha^k (L +D)^{-1} \left[  X^T X \mathbf s^{k-1} - X^T \mathbf y  \right].
\end{aligned}
\end{equation}

When $w=1$ and $\alpha^k=1$, both SOR and CD+SRRC reduce to the traditional coordinate descent.
Equation~\eqref{eq:sor:formulation} and Equation~\eqref{eq:srrc:iteration} share the following two similaries:
1) both make use of the gradient in the recursive iterations in that $X^TX \bm \beta^{k-1}  - X^T \mathbf y$ is the gradient of $\frac{1}{2} \| X \bm \beta - \mathbf y\|_2^2$ at
$\bm \beta^{k-1}$ and $X^TX \mathbf s^{k-1}  - X^T \mathbf y$ is the gradient of $\frac{1}{2} \| X \mathbf s^{k-1} - \mathbf y\|_2^2$ at
$\mathbf s^{k-1}$, and 2) both use a precondition matrix in that SOR uses $(wL+D)$ whereas SRRC uses
$(L +D)$. A key difference is that the precondition matrix used in SRRC is parameter-free whereas the one used in SOR has a parameter.
As a result, we can perform an inexpensive univariate search to find the optimal $\alpha^k$ used in SRRC whereas it is 
usually expensive for SOR to search for 
an optimal $w$ in the same way as Equation~\eqref{eq:one:d:search}.

When the design matrix has some special structures, it has been shown in~\cite{Yong_D:1950} that the optimal value of $w$ can be found for SOR.
However, for the general design matrix $X$, it is hard to obtain the optimal $w$ used for SOR. 
This might be a major reason that SOR is not widely used in solving Lasso with coordinate descent.
For our proposed SRRC, the criterion in Equation~\eqref{eq:one:d:search} enables us to adaptively set 
the refinement factor $\alpha^k$.

\subsection{Relationship between SRRT and\\ the Nesterov's Method}

The SRRT scheme presented in Figure~\ref{fig:sort:diagram} is similar to the Nesterov's method in that both make use of a search point in the iterations.
In addition, both set the search point using
\begin{equation}
   \mathbf s^k = (1-\alpha^k) \bm \beta^{k-1} + \alpha^k \bm \beta^k.
\end{equation}
However, the key difference is that the $\alpha^k$ used in the Nesterov's method is predefined according to a specified formula, whereas the $\alpha^k$ used in
SRRT is set to optimize the objective function as shown in Equation~\eqref{eq:one:d:search}.
Note that if the Nesterov's method sets the $\alpha^k$ to optimize the objective function, it reduces
to the traditional steepest descent method thus the good acceleration property of the Nesterov's method is gone.

%% file: experiment.tex
\begin{table}
	\caption{Performance for the synthetic data sets. The results are averaged over 10 runs. 
	The sparsity is defined as the number of zeros in the solution divided by the total number of variables $p$.}
	\label{tab:SyntheticResults}
\begin{center}
\begin{small}
		\begin{tabular}{l|lllll}
		\hline
		 data  size   &  $\lambda$    
		                                                 &  CD      & CD+SRRC & CD+SRRT & sparsity \\
		\hline
		\multirow{2}{*}{$n=500$}          & 0.5          &  10.0    & 8.8       & 9.2       & 0.9395 \\
		\cline{2-6}
		                                  & 0.1          &  151.7   & 74.7      & 59.5      & 0.6406 \\
		\cline{2-6}
		\multirow{2}{*}{$p=1000$}         & 0.05         &  463.2   & 179.0     & 109.1     & 0.5763 \\
		\cline{2-6}
		                                  & 0.01         &  4132.7  & 1419.4    & 326.1     & 0.5146\\
		\hline
		\hline
		\multirow{2}{*}{$n=1000$}         & 0.5          &  7.9     & 7.7       & 7.7       & 0.9397  \\
		\cline{2-6}
		                                  & 0.1          &  54.9    & 31.7      & 29.0      & 0.473  \\
		\cline{2-6}
		\multirow{2}{*}{$p=1000$}         & 0.05         &  125.4   & 59.9      & 47.7      & 0.3126\\
		\cline{2-6}
		                                  & 0.01         &  748.0   & 293.4     & 128.9     & 0.118\\
		\hline
		\hline
		\multirow{2}{*}{$n=1000$}         & 0.5          & 7.9      & 7.7       & 7.8      & 0.8856 \\
		\cline{2-6}
		                                  & 0.1          & 26.3     & 17.3      & 17.1     & 0.3102 \\
		\cline{2-6}
		\multirow{2}{*}{$p=500$}          & 0.05         & 35.5     & 21.3      & 20.2     & 0.1678  \\
		\cline{2-6}
		                                  & 0.01         & 47.9     & 26.3      & 25.1     & 0.0382 \\
		\hline
		\end{tabular}
\end{small}
\end{center}
\end{table}

\begin{table}
	\caption{Performance for the real data sets. 
	The sparsity is defined as the number of zeros in the solution divided by the total number of variables $p$.}
	\label{tab:RealResults}
\begin{center}
\begin{small}
		\begin{tabular}{l|lllll}
		\hline
		 data  size   &  $\frac{\lambda}{\|X^T\mathbf y\|_{\infty}}$    
		                                                 &  CD       & CD+SRRC & CD+SRRT & sparsity \\
		\hline
		\multirow{2}{*}{$n=38$}        & 0.5             &  122      & 68        & 84         & 0.9982\\
		\cline{2-6}
		                               & 0.1             &  155      & 90        &  103       & 0.9964 \\
		\cline{2-6}
		\multirow{2}{*}{$p=7129$}
						                       & 0.05            &  254      & 119       &  127       & 0.9961 \\
		\cline{2-6}
		                               & 0.01            &  2053     & 424       &  343       & 0.9948 \\
		\hline
		\hline
		\multirow{2}{*}{$n=62$}        & 0.5             &  31       & 21        & 24         & 0.9975\\
		\cline{2-6}
		                               & 0.1             &  157      & 68        & 78         & 0.9840 \\
		\cline{2-6}
		\multirow{2}{*}{$p=2000$}
						                       & 0.05            &  308      & 115       & 118        & 0.9775 \\
		\cline{2-6}
		                               & 0.01            &  2766     & 929       & 375        & 0.9715 \\
		\hline
		\hline
		\multirow{2}{*}{$n=6000$}      & 0.5             &  26      & 16         & 11          & 0.9994\\
		\cline{2-6}
		                               & 0.1             &  180     & 103        & 108         & 0.9932\\
		\cline{2-6}
		\multirow{2}{*}{$p=5000$}
						                       & 0.05            &  823     & 337       &  432         & 0.9876\\
		\cline{2-6}
		                               & 0.01            &  4621    &  1387     &  1368        & 0.8340\\
		\hline
		\end{tabular}
\end{small}
\end{center}
\end{table}

In this section, we report experimental results for synthetic and real data sets, studying the number of iterations
of CD, CD+SRRC and CD+SRRT for solving Lasso. The consumed computational time is proportional to the number of iterations.

\vspace{0.1in}

\noindent  \textbf{Synthetic Data Sets }
We generate the synthetic data as follows. The entries in the $n \times p$ design matrix $X$ and
the $n \times 1$ response $\mathbf y$ are drawn from a Gaussian distribution.
We try the following three settings of $n$ and $p$: 1) $n=500, p=1000$, 2) $n=1000,p=1000$, and 3) $n=1000, p=500$. 

\vspace{0.1in}

\noindent  \textbf{Real Data Sets }
We make use of the following three real data sets provided in~\cite{libsvmData}: leukemia, colon, and gisette.
The leukemia data set has $n=38$ samples and $p=7129$ variables.
The colon data set has $n=62$ samples and $p=2000$ variables.
The gisette data set has $n=6000$ samples and $p=5000$ variables.

\begin{figure*}
  \centering
	 synthetic ($n=500,p=1000$)  \hspace{0.15 \columnwidth} synthetic ($n=1000,p=1000$) 
	\includegraphics[width=0.45\columnwidth]{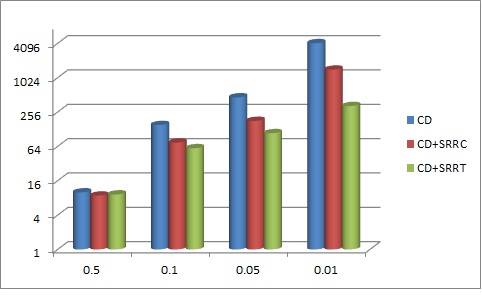}
	\includegraphics[width=0.45\columnwidth]{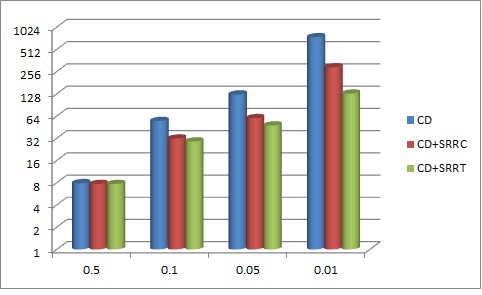}\\
	\vspace{0.2in}
	 synthetic ($n=1000,p=500$) \hspace{0.15 \columnwidth} leukemia ($n=38,p=7129$) \\
	\includegraphics[width=0.45\columnwidth]{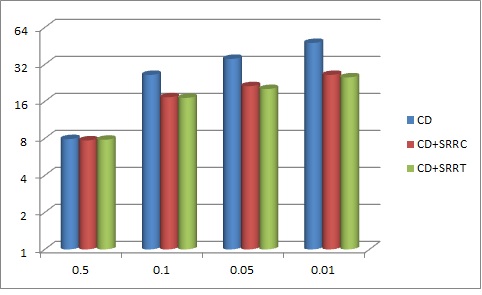}	
	\includegraphics[width=0.45\columnwidth]{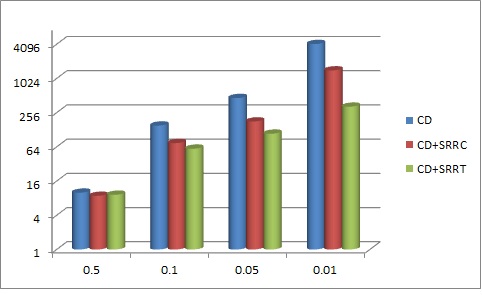}\\
	\vspace{0.2in}
	colon ($n=62,p=2000$)  \hspace{0.15 \columnwidth} gisette ($n=6000,p=5000$)\\
	\includegraphics[width=0.45\columnwidth]{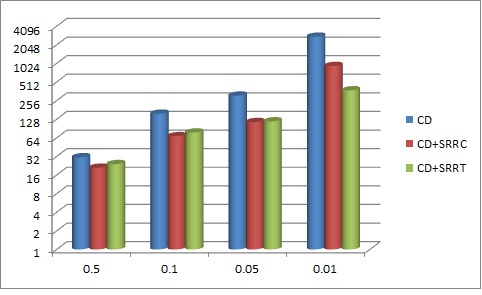}
	\includegraphics[width=0.45\columnwidth]{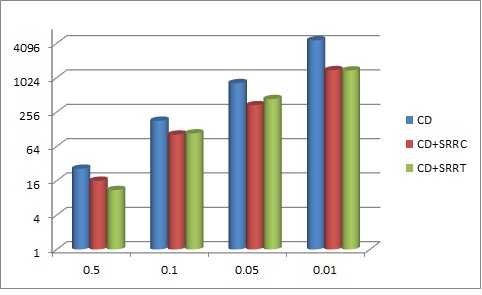}
	\caption{Number of iterations used by CD, CD+SRRC, and CD+SRRT on synthetic and real data sets. 
	For all the plots, the x-axis corresponds to the regularization parameter $r = \frac{\lambda}{\|X^T \mathbf y\|_{\infty}}$
	and the y-axis denotes the number of iterations in a logarithmic scale by different approaches.}
	\label{fig:result}
\end{figure*}

\vspace{0.1in}

\noindent  \textbf{Experimental Settings }
For the value of the regularization parameter, we try $\lambda =r \|X^T \mathbf y\|_{\infty}$, where $r=0.5,0.1,0.05,0.01$.
For the synthetic data sets, the reported results are averaged over 10 runs.
For a particular regularization parameter, we first run CD in Algorithm~\ref{algorithm:traditional} until 
$\|\bm \beta^k -\bm \beta^{k-1}\|_2 \le 10^{-6}$, and then run CD+SRRC and CD+SRRT until the obtained 
objective function value is less than or equal to the one obtained by CD.

\vspace{0.1in}

\noindent  \textbf{Results } Table~\ref{tab:SyntheticResults} and Table~\ref{tab:RealResults}
show the results for the synthetic and real data sets, respectively. The last column
of each table shows the sparsity of the obtained Lasso solution, which is defined as the number of
zero entries in the solution divided by the number of variables $p$. 
Figure~\ref{fig:result} visualizes the results in these two tables.
We can see that when the solution is 
very sparse (for example, $\lambda = 0.5 \|X^T \mathbf y\|_{\infty}$), the proposed CD+SRRC and CD+SRRT
consume comparable number of iterations to the traditional CD. The reason is that the optimal
refinement factor computed by SRR in Equation~\eqref{eq:one:d:search} is equal to or close to 1,
and thus CD+SRRC and CD+SRRT is very close to the traditional CD.
Note that a regularization parameter $\lambda = 0.5 \|X^T \mathbf y\|_{\infty}$ is usually too large
for practical applications because it selects too few variables, and we usually need to try a smaller
$\lambda = r \|X^T \mathbf y\|_{\infty}$ for example, $r=0.01$. It can be observed that the proposed
CD+SRRC and CD+SRRT requires much fewer iterations, especially for smaller regularization parameters.

%% file: conclusion.tex
In this paper, we propose a novel technique called successive ray refinement. 
Our proposed SRR is motivated by an interesting ray-continuation property on the coordinate descent iterations:
for a particular coordinate, the value obtained in the next iteration almost always
lies on a ray that starts at its previous iteration and passes through the current iteration.
We propose two schemes for SRR and apply them to solving Lasso with coordinate descent.
Empirical results for real and synthetic data sets show that the proposed SRR can significantly reduce the number of coordinate descent iterations, especially when the regularization parameter is small.

We have established the convergence of CD+SRR, and it is interesting to study the convergence rate.
We focus on a least squares loss function in~\eqref{eq:lasso:problem}, and
we plan to apply the SRR technique to solving the generalized linear models.
We compute the refinement factor as an optimal solution to Equation~\eqref{eq:one:d:search}, and
we plan to obtain the refinement factor as an approximate solution, especially in the case of generalized linear models.